\documentclass[letterpaper, 10 pt, conference]{ieeeconf}  
\IEEEoverridecommandlockouts
\usepackage{graphicx} 
\usepackage{multicol}
\usepackage{relsize}
\usepackage{times} 
\usepackage{amsmath}
\usepackage{amssymb}
\usepackage{algorithm,algorithmic}
\usepackage{textcomp}
\usepackage[flushleft]{threeparttable}
\usepackage{cite}
\usepackage[bookmarks=true]{hyperref}
\usepackage[hang,flushmargin]{footmisc}

\title{\LARGE \bf
	Data-Driven Gait Segmentation for Walking Assistance in a Lower-Limb Assistive Device}
 \author{Aleksandra Kalinowska$^{1}$, Thomas A. Berrueta$^{1}$, Adam Zoss$^{3}$, and Todd Murphey$^{1,2}$%
 \thanks{\quad *This work was supported by the National Science Foundation (NSF) under grant CBET-1637764. Any opinions, findings, and conclusions or recommendations expressed in this material are those of the authors and do not necessarily reflect the views of the NSF. }
 \thanks{\quad $^{1}$Mechanical Engineering, Northwestern University, Evanston, IL; $^{2}$Physical Therapy and Human Movement Science, Northwestern University, Chicago, IL; $^{3}$Ekso Bionics, Richmond, CA.}
}

\begin{document}
	\maketitle
	\thispagestyle{empty}
	\pagestyle{empty}
	
	\begin{abstract}
	    Hybrid systems, such as bipedal walkers, are challenging to control because of discontinuities in their nonlinear dynamics. Little can be predicted about the systems' evolution without modeling the guard conditions that govern transitions between hybrid modes, so even systems with reliable state sensing can be difficult to control. We propose an algorithm that allows for determining the hybrid mode of a system in real-time using data-driven analysis. The algorithm is used with data-driven dynamics identification to enable model predictive control based entirely on data. Two examples---a simulated hopper and experimental data from a bipedal walker---are used. In the context of the first example, we are able to closely approximate the dynamics of a hybrid SLIP model and then successfully use them for control in simulation. In the second example, we demonstrate gait partitioning of human walking data, accurately differentiating between stance and swing, as well as selected subphases of swing. We identify contact events, such as heel strike and toe-off, without a contact sensor using only kinematics data from the knee and hip joints, which could be particularly useful in providing online assistance during walking. Our algorithm does not assume a predefined gait structure or gait phase transitions, lending itself to segmentation of both healthy and pathological gaits. With this flexibility, impairment-specific rehabilitation strategies or assistance could be designed.
	\end{abstract}

	\section{Introduction}
	In order to implement predictive control on an autonomous system or to provide robotic assistance to a human, we require the ability to identify system dynamics. Lots of work has been done in this area, allowing one to learn the dynamics of an autonomous robot~\cite{koopman_ian,deisenroth2015,pan2012,mpc_neuralnets,modellearning_review} or a joint human-machine system~\cite{koopman_alex} from collected state data.
	
	Hybrid systems are more challenging. Even if we have reliable sensors to real-time detect the switching times of the system dynamics, we require a closed-form mapping from states to hybrid mode to implement any form of predictive control. And, oftentimes, one would prefer to avoid installing supplementary sensors altogether, because of the cost, unreliability or inconvenience of additional hardware. 
	
	Much work in this domain has been done in the context of gait analysis and gait phase identification. Real-time, closed-form expressions for distinguishing between hybrid dynamic modes can be obtained using supervised machine learning techniques, such as neural networks~\cite{imu_neuralnet,mlp_joint_angles}, Hidden Markov Models~\cite{HMM_gaitpartitioning}, or Gaussian Mixture Models~\cite{GMM_gaitpartitioning} from selected sensory information, such as ground contact forces, joint positions, inertial data, or muscle signals. Most of these studies rely heavily on ankle data for phase identification, which is often not present in lower-limb assistive devices, such as the one used in this study. Moreover, work in this area often requires assumptions to be made about the number of dynamic modes, phase transition times, or phase durations, limiting the algorithms' ability to adequately partition abnormal gait. Finally, to the authors' best knowledge, no gait partitioning strategies have been shown to have the potential to generate control for assistance using a model predictive controller in real-time. 
	
	Here, we propose a novel algorithm that synthesizes a closed-form expression for hybrid mode switching conditions, allowing for real-time switching time predictions. For a simple system---a simulated one-legged hopper---we show that the data-derived guard conditions can be employed for control. For a bipedal walker (a human in an Ekso Bionics\textsuperscript\textregistered~exoskeleton), we demonstrate accurate gait segmentation using data only from knee and hip joints. We validate our gait partitioning by comparing it to pressure data from heel and toe contact sensors and demonstrate the ability to reliably predict heel strike and toe-off events without use of impact sensors. We do not pre-label transitions in our training data or pre-define the expected number of phases, which allows us to identify a range of recurring movement patterns in the gait cycle and shows promise for meaningful partitioning of abnormal walking.   
	
	This paper lays the groundwork for applications of real-time assistance. In Section \ref{methods}, we give a detailed description of our algorithmic approach and methods used. In Sections \ref{slip} and \ref{sec: exo}, we validate our methodology through two examples: control of a simulated hopper without any \textit{a priori} knowledge of its dynamics or guard conditions and gait partitioning of experimental data from a bipedal walker. Finally, we conclude with a discussion of the results and opportunities for future work. 
	
	\section{Methods}
	\label{methods}
	
	Our procedure for approximating the dynamics of a hybrid system for control is divided into two parts: synthesizing switching conditions between dynamically distinct modes, and estimating the continuous dynamics of each mode from data. For the first part, we use Nonparametric Clustering of Dynamics (NCD), where we locally approximate system dynamics at each point in time, apply a nonparametric clustering algorithm onto those dynamical models, and train a classifier to obtain a mapping from system states to dynamically distinct modes. We later use this mapping to identify a system's hybrid mode in real-time. For the second part, we use Koopman operators~\cite{koopman_ian}, which allow us to generate an approximation of the within-mode continuous dynamics. Both of these approaches are described in more detail in the subsections below. 

	\subsection{Nonparametric Clustering of Dynamics}
	\label{sec: NCD}

	Hybrid dynamical systems evolve according to distinct dynamics in different regions of the state-space manifold and as a result are often represented as finite automata with discrete nodes~\cite{henzinger1996}. Here we propose NCD---a system identification algorithm that allows us to synthesize finite representations of dynamical systems from data. Since hybrid systems express distinct dynamics based on state-space dependencies, local estimates of the dynamics differ about the boundaries specified by the system's guard equations. NCD generates partitions in the state-space manifold, where local estimates of the system dynamics differ. It numerically approximates boundaries on the manifold that delineate which mode governs the system at a given point and, as a result, identifies transitions between modes in hybrid systems, otherwise defined by guard equations.
	
	Given a dataset representative of all dynamic modes in a hybrid system, NCD first segments the dataset into subsets and then generates local estimates of the system dynamics from each subset. Subsets can have various temporal lengths, they can be independent or overlap; however, they should always be continuous in time and have the same dimensionality. While any function approximation technique that produces a numerical closed-form model of the system dynamics can be used, we use generalized linear regression to generate Koopman operators (see Section \ref{sec:koop}), because their representation makes them easy to use in control settings \cite{williams2016}. In Koopman operators, states $x \in \mathbb{R}^n$ are lifted into a higher dimensional space through a nonlinear transformation of the original state-space by basis functions $\Psi(x) = [\psi_1(x), ..., \psi_N(x)]^T \in \mathbb{R}^N \ s.t. \ \psi_i(x) \in \mathbb{R}$. 
	
	The collection of local dynamic models is then made into a list, where we apply unsupervised learning techniques to divide the list into classes of distinct dynamic models based on a distance metric. For models generated by generalized regression, the list contains the weight matrices from each local model, $L = [W_0, .., W_{S-1}]$ from $S$ subsets of data. For other kinds of parametrized models, the list may consist of parameter vectors. If the number of hybrid modes is known ahead of time, one can apply parametric clustering techniques and prespecify the expected number of clusters to form. Nonparametric clustering techniques are useful for analyzing systems where the partitions are not as well-defined. We apply Hierarchical Density-Based Clustering for Applications with Noise (HDBSCAN) to perform nonparametric clustering~\cite{hdbscan}.
	
	Once all dynamic models in the list have been assigned a class label by the clustering algorithm, one can extend the class labels to each model's corresponding subset of data. To map the class labels onto the state-space, we train a Support 

		\noindent\begin{minipage}{\columnwidth}
    \renewcommand\footnoterule{} 
	\begin{algorithm}[H]
		\caption{Nonparametric Clustering of Dynamics (NCD)}
		\noindent\footnotetext{*Here, we use generalized linear regression to generate Koopman operators, described in Section~\ref{sec:koop}.}
		\begin{algorithmic}[1]
			\renewcommand{\algorithmicrequire}{\textbf{Input:}}
			\renewcommand{\algorithmicensure}{\textbf{Procedure:}}
			\REQUIRE Dataset $X = [x_0,...,x_M]$, function approximation technique with closed-form model*.
			\ENSURE 
			\STATE Split $X$ into $S$ subsets
			\STATE Estimate system dynamics locally for each subset of the dataset $X_{a:b} = \{X[a],...,X[b]\}$ using $W = FunctionApprox(X_{a:b})$
			\STATE Construct list of dynamic models $L = [W_0,...,W_{S-1}]$
			\STATE Apply nonparametric clustering to $L$ and label all $W_i$'s with one of $B$ discerned classes $\{C_0,...,C_{B-1}\}$
			\STATE Label all points in $X$ with the label $l \in \{0,...,B-1\}$ of the subset they were applied to
			\STATE Train an SVM, $\Phi(x)$, to project class labels directly onto the state-space
			\renewcommand{\algorithmicensure}{\textbf{Return:}}
			\ENSURE Trained SVM indicator function $\Phi(x)$
			\label{alg:NCD}
		\end{algorithmic}
	\end{algorithm}
	\vspace{-0.7cm}
	\renewcommand\footnoterule{}
    \end{minipage}
    \vspace{0.35cm}	
	
	\noindent Vector Machine (SVM) on the labeled data to generate an indicator function $\Phi(x)$~\cite{machinelearning}. In particular, we generate an indicator function based on nonlinear transformations of the data, $\Phi(\Psi(x))=i,$ where $i \in \{0,..., B-1\}$ of $B$ discerned dynamic modes. The indicator function specifies what mode the system is currently in, similar to the guard equations in hybrid systems. Algorithm 1 summarizes the NCD procedure for generating data-driven hybrid mode transition boundaries.

	The indicator function gives a closed-form estimate of the regions drawn by the guard equation boundaries, which allows one to predict switching times in hybrid systems via model predictive control. If the hybrid mode dynamics are not known \textit{a priori}, one can generate data-driven models of each hybrid mode by segmenting the dataset according to the partitions designated by NCD and learning a model for each dynamic subsystem.

	\subsection{Koopman Operators}
	\label{sec:koop}
	
	Koopman operators have been shown to be effective in modeling observable dynamical systems~\cite{koopman_ian}. Formally, Koopman operators describe the time-evolution of dynamical systems in an infinite-dimensional function space~\cite{koopman,budisic2012}. While the infinite-dimensional Koopman operator is valuable as a theoretical construct, it is impractical for numerical applications. Through generalized linear regression, one can synthesize finite-dimensional approximations of the Koopman operator by considering nonlinear basis functions of state $\Psi(x) = [\psi_1(x), ..., \psi_N(x)]^T$ and their evolution in time~\cite{williams2016}. The regression generates an operator $K$ that minimizes the residual $r(x_k)$ in $\Psi(x_{k+1}) = K \Psi(x_k) + r(x_k)$ through the least-squares optimization
	
	\begin{equation}
	\underset{K}{\text{min}} \ \frac{1}{2}\mathlarger{\sum}_{k=1}^{M-1}||\Psi(x_{k+1})-K\Psi(x_k)||^2.
	\label{eq:koop_opt}
	\end{equation}
	
	The optimization has closed-form solution $K = AG^\dagger$, where $\dagger$ denotes the Moore-Penrose pseudoinverse, and the matrix components $A$ and $G$ are
	\vspace{-0.15cm}
	\begin{eqnarray}
	G &=& \mathlarger{\frac{1}{M}}\mathlarger{\sum}_{k=1}^{M-1} \Psi(x_k)\Psi(x_k)^T \nonumber \\
	A &=& \mathlarger{\frac{1}{M}}\mathlarger{\sum}_{k=1}^{M-1} \Psi(x_{k+1})\Psi(x_k)^T.
	\label{eq:opt_mats}
	\end{eqnarray}
	\vspace{-0.2cm}
	
	We can apply NCD in conjunction with finite-dimensional Koopman operators to generate data-driven estimates of the state-space boundaries designated by the guard equations and to synthesize the dynamical models of each hybrid mode.

	\section{Example 1: Control of a Simulated SLIP} \label{slip}
	
	The spring-loaded inverted pendulum (SLIP), often used as a simplified model of human running, is an example of a hybrid system with known dynamics. Although governed by relatively simple equations, it is unstable if left unassisted. Here, we use a simulated SLIP to demonstrate the joint capabilities of NCD and Koopman operators for predicting hybrid mode switching and identifying bimodal dynamics for control. In simulation, we successfully use the data-derived approximation of the SLIP's hybrid dynamics to generate forward motion using model predictive control. 
	
	\subsection{Switching Time and Dynamics Identification}
	
	For all simulations, we use a 2D SLIP model described by a state vector $x = [x_m, \dot{x}_m, z_m, \dot{z}_m, x_t]$, where $x_m$ and $z_m$ are the coordinates of the mass, and $x_t$ is the coordinate of the toe, and a control vector $u = [u_{stance},u_{flight}]$, where $u_{stance}$ is the leg thrust applied during stance and $u_{flight}$ is the toe velocity control applied during flight. SLIP dynamics as described in~\cite{ola_RSS} are used. We begin by generating 30 seconds of training data using a simulation of the SLIP hopper controlled by a model predictive controller with knowledge of the correct SLIP dynamics.  
	
\begin{figure}[t]
		\centering
		\includegraphics[width=0.85\columnwidth]{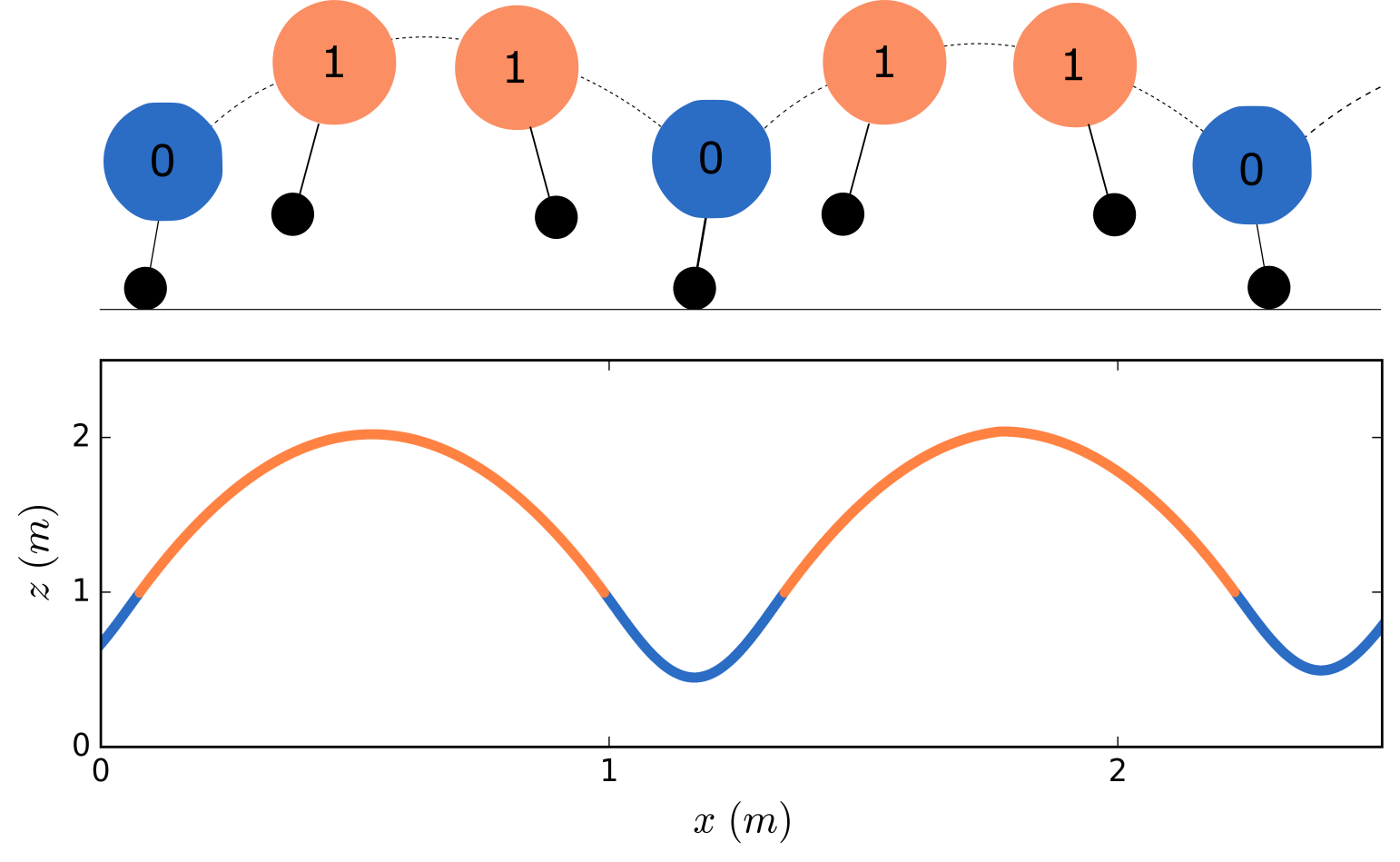}
		\caption{SLIP trajectory segmented according to an NCD-generated mapping. The data-driven indicator function is able to detect the SLIP's hybrid mode with $>99\%$ accuracy for hopping on flat ground.}
		\label{fig:segmentation}
	\end{figure}

	We proceed with system identification by employing NCD, described in Section~\ref{sec: NCD}. The algorithm generates an indicator function that maps from states to SLIP hybrid modes. We verify the mapping by directly comparing against the solution of the analytical guard equation~\cite{ola_RSS} and find that the data-driven indicator function is able to detect hybrid modes with near-perfect accuracy for two tested trajectories---it is 100\% accurate for constant-velocity forward hopping and 99.5\% accurate for varying-velocity hopping with directional changes.  A fragment of the constant-velocity trajectory color-coded according to the SLIP's current hybrid mode is shown in Fig.~\ref{fig:segmentation}. 
	
	Finally, using the NCD-generated mapping, we synthesize two separate Koopman operators for the SLIP's flight and stance modes. This obtained, entirely data-derived representation of the SLIP dynamics is then used for control. 
	
	\begin{figure}[t]
		\centering
		\includegraphics[width=0.85\columnwidth]{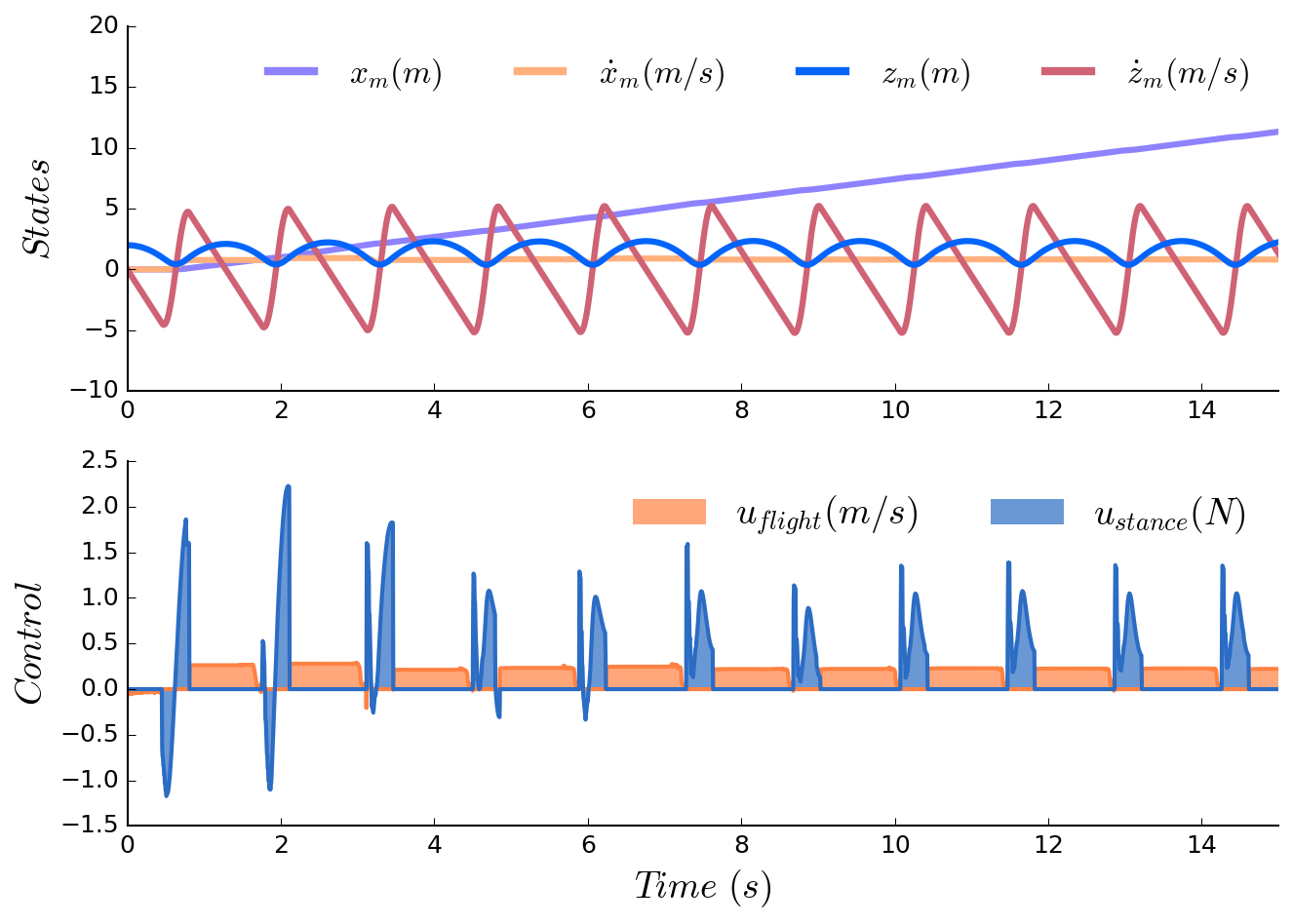}
		\caption{State trajectories and control history from a SLIP model simulation---the guard equation and system dynamics used in the simulation were learned solely from data. Note that the controller is able to keep the SLIP upright and moving forward at an average speed of $0.37m/s$, close to the desired $0.4m/s$.}
		\label{fig: SLIP_control}
	\end{figure}
	
	\subsection{Model Predictive Control}
	
	A model predictive controller (MPC) similar to \cite{alex_SAC} is used. However, any model predictive controller that is capable of completing the task can be used. For the MPC, we define an objective function of the form
	\begin{equation}
	J = \frac{1}{2} \int_{t_o}^{t_f} \lVert x(t) - x_d(t)\rVert_{Q}^2 + \lVert u(t) \rVert_{R}^2  \ dt,
	\end{equation}
	with $Q \ge 0$ and $R \ge 0$ being cost metrics on state error and control effort, respectively, and $x_d(t)$ being the desired trajectory. 

	We define the system as
\vspace{-0.5cm} 
\begin{center}
\[ \Psi(x_{k+1}) =  \left\{ \begin{array}{ll}
  			K_{stance} \Psi(x_k) & \Phi(x_k)=0 \\
  			K_{flight} \Psi(x_k) & \Phi(x_k)=1, \\
		\end{array}  \right. \]
\end{center}
where $K_{stance}$ and $K_{flight}$ are the two learned Koopman operators and $\Phi(x)$ is the learned indicator function that can take values $\{0,1\}$ corresponding to stance and flight, respectively. 
	
	For an example trial, we run a simulation with a desired trajectory $x_d(t) = [0, 0.4, 1.6, 0, 0]$, and a diagonal $Q$ matrix with $Q_{diag}=[0,50,100,0,0]$. As such, the controller tries to maintain the SLIP center of mass at a height of $1.6m$ and a forward velocity of $0.4m/s$. In this trial, the SLIP starts at a height $z_m=2m$ and no forward velocity ($\dot{x}_m=0m/s$). As shown in Fig.~\ref{fig: SLIP_control}, the controller is able to keep the SLIP upright and moving forward at a velocity of $0.37m/s$, close to the desired $0.4m/s$, while having knowledge solely of the data-driven dynamics.
	
	\section{Example 2: Gait Partitioning for a Biped} \label{sec: exo}
	
	Gait partitioning is an area of interest due to its promising applications in improving control of lower-limb assistive devices as well as in generating individually tailored physical therapies. Here, we segment the gait cycle into phases using the NCD algorithm, described in Section~\ref{sec: NCD}, and interpret obtained mode transitions based on established gait events. With control generation in mind, we are particularly interested in determining impact events, such as heel strike and toe-off, because they mark transitions between dynamically distinct modes. We validate the accuracy and latency of our predictions against external pressure sensors. Our results demonstrate successful gait partitioning for healthy flat-ground walking.
	
	\subsection{Walking Data Acquisition}
	\label{sec: ekso_data}
	
	\begin{figure}[t]
		\centering
		\includegraphics[width=0.795\columnwidth]{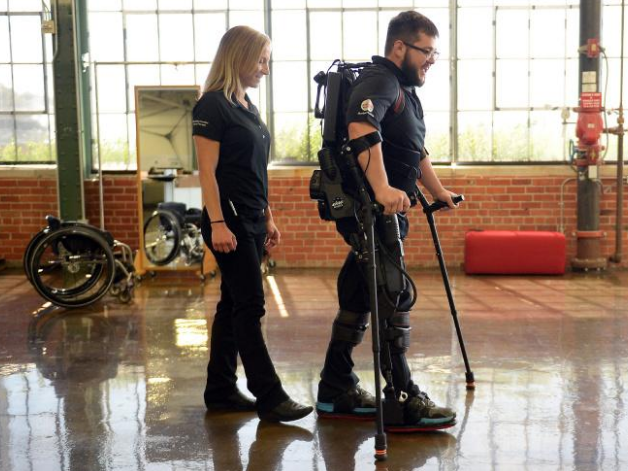}
		\vspace{-0.2cm}
		\caption{Ekso Bionics exoskeleton, EksoGT\texttrademark, used for data collection. We use two sets of its sensors in this study: hip and knee encoders to generate kinematics-based dynamical models using the NCD algorithm, and foot-mounted pressure sensors at the heel and toe to validate our models against ground-contact events after partitioning.}
		\label{fig: ekso}
		\vspace{-0.3cm}
	\end{figure}
	
	Data was collected using EksoGT\texttrademark---a robotic exoskeleton from Ekso Bionics, Richmond, CA, USA, visible in Fig.~\ref{fig: ekso}. When not actively in assistance mode, the device offers freedom to move in the sagittal plane, and to a limited extent in the frontal plane. It provides assistance solely in the sagittal plane. Both knee and hip joints can be used for assistance, where angular position and angular velocity can be measured at $500Hz$ by encoders in all four joints. The ankle joints are passive and no sensory data is available. 
	
	For the purposes of this study, $1.5$ minutes of data were collected of straight flat-ground walking from one healthy subject with previous experience walking in the exoskeleton (the third author of this paper). No assistance or resistance was provided to the wearer through motor activity; any perceptible resistance was passive from the mechanical structure of the device. A total of sixteen variables were recorded. Twelve of them (right/left knee angles, right/left knee angular velocities, right/left hip angles, right/left hip angular velocities, right/left knee motor currents, and right/left hip motor currents) were used for analysis. The additional four variables (right/left toe sensors and right/left heel sensors) were excluded from analysis and used solely for validation and verification of NCD-generated gait partitions.
	
	During our data analysis, we use $1.33$ out of the recorded $1.5$ minutes of straight overground walking---we discard the first $10$ seconds due to irregularities introduced by gait initiation. We train our data on a randomly selected $20$-second sample from the trimmed $1.33$-minute window and test the segmentations on the whole trimmed dataset.	
	
	\begin{figure}[t]
		\centering
		\includegraphics[width=0.85\columnwidth]{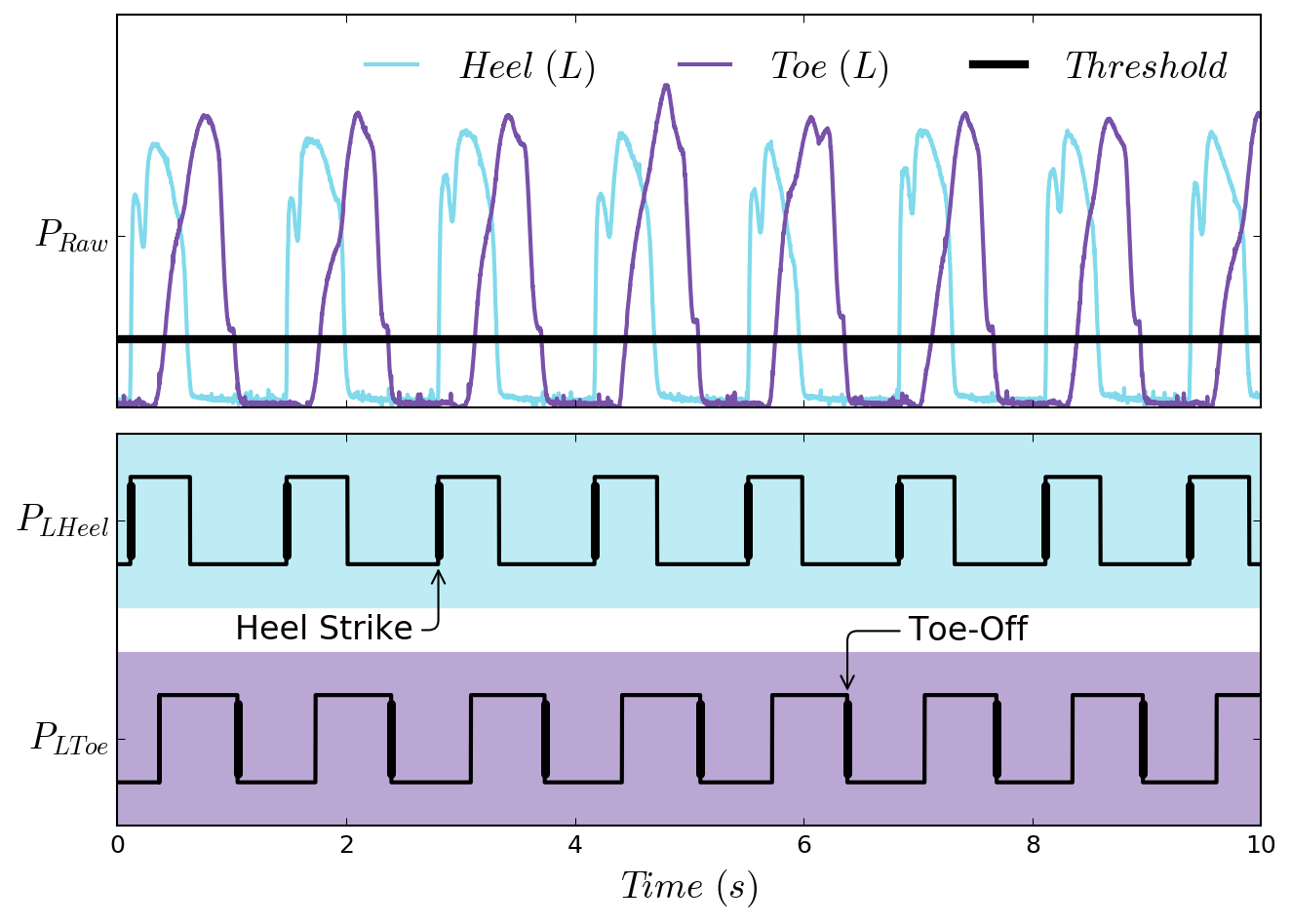}
		\vspace{-0.3cm}
		\caption{Example signals from two foot-mounted pressure sensors of the exoskeleton in Fig.~\ref{fig: ekso}. The top plot shows raw signals from the left heel and left toe sensors for several gait cyclces. These raw signals are thresholded to generate binary digital signals visible underneath, which allow us to detect gait events such as heel strikes and toe-offs on rising edges of the heel signal and falling edges of the toe signal, respectively. }
		\label{fig: sensors}
		\vspace{-0.3cm}
	\end{figure}	
	
	\begin{table*}[t]
	\begin{center}
		\begin{threeparttable}
			\caption{Gait Decompositions}
			\label{table: summary}
			\begin{center}
				\begin{tabular}{c|c|c|c}
					\hline
					& \textbf{2-mode decomposition} & \textbf{4-mode decomposition} & \textbf{6-mode decomposition} \\
					\hline
					\textbf{Phases} & right/left step & right/left pre-swing+swing and stance & right/left initial swing, terminal swing, and stance\\ 
					\hline
					\textbf{Events} & heel strike & heel strike and onset of knee buckling & heel strike, toe-off, and foot clearance at peak swing \\
					\hline
					\textbf{False positives*} & 0\% & 2\% & 8\% \\
					\hline
					\textbf{False negatives*} & 0\% & 4\% & 1\% \\
					\hline
					\textbf{Average offset*} & $6.1ms \pm 5.2ms$ & $13.0ms \pm 8.0ms$ & $20.0ms \pm 9.8ms$ \\
					\hline
				\end{tabular}
				\begin{tablenotes}
					\scriptsize 
					\item *Classification rate and average offsets were calculated only for predictions of impact events, specifically heel strikes and toe-offs. These events were verified against signal from external pressure sensors not used in generating the data-driven gait partitions. A total of 220 such events were recorded during the tested 1.33-minute time window. Average offset was calculated only based on correctly identified transitions.  
				\end{tablenotes}
				\vspace{-0.5cm}
			\end{center}
		\end{threeparttable}
	\end{center}
	\end{table*}

	\subsection{Ground Truth for Gait Partitions}
	\label{sec: validation}
	
	Gait cycles are generally defined from one foot strike to the subsequent foot strike on the same side. Clinically, they are often partitioned separately for each leg based on functionally critical events for that leg~\cite{gaitanalysis}. For this study, we are interested in generating gait partitions that capture critical changes in the behavior of both legs. We do not impose symmetry, but we choose states and basis functions symmetrically for right and left legs to allow the algorithm to remain equally sensitive to recurring patterns on both sides. Moreover, we look for gait events representing contact with the ground, specifically heel strike and toe-off, because these impact events indicate transitions between dynamically distinct modes that are important for generating control in a robotic assistive device. As a tertiary objective, we are interested in sub-dividing swing, because the majority of active assistance during walking takes place during the swing phase. To generate data-driven approximations for transitions between the described dynamically distinct gait phases, we use the proposed algorithm---NCD. 
	
			\begin{figure}[!t]
		\centering
		\includegraphics[width=0.85\columnwidth]{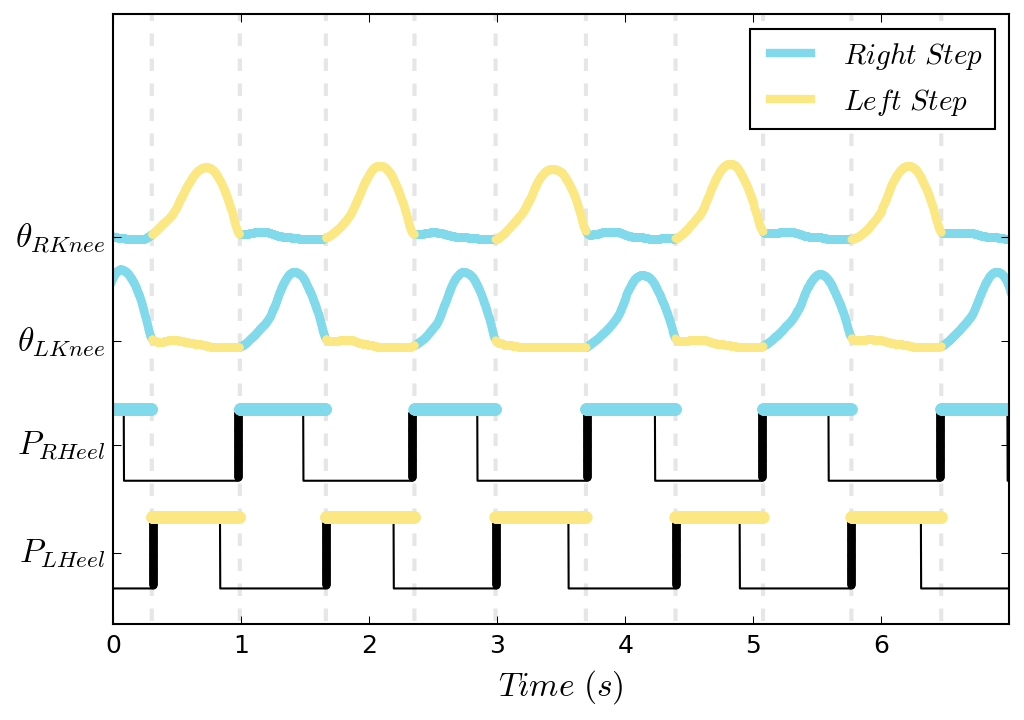}
		\vspace{-0.4cm}
		\caption{Two-phase segmentation of the gait cycle plotted against heel pressure sensors. Dotted gray lines depict NCD predictions of heel strikes, while bold black segments show ground truth. Identified phases are superimposed onto knee trajectories depicting their relationship to the gait cycle. We detect heel strikes without misclassifications and an average offset of $6.1ms \pm 5.2ms$. }
		\label{fig: 2phase_detection}
		\vspace{-0.3cm}
	\end{figure}
	
	In order to validate NCD-generated gait partitions, we utilize foot-mounted pressure sensors at the heel and toe. We collect analog signals from the pressure sensors and threshold them to obtain binary readings of whether the heel and toe are in contact with the ground. These processed sensor readings allow us to directly record heel strikes and toe-offs, establishing a notion of ground truth for transitions between stance and swing phases, as demonstrated in Fig.~\ref{fig: sensors}. As a result, for each gait decomposition, we can find mode transitions that correspond to these ground-contact events, and measure the offset between the prediction of the event and the ground truth from the pressure sensors. The offset measurement is dependent on NCD reliably recognizing specific mode transitions and thus conveys both the precision and accuracy of event detection. We report the offset, or latency, as our validation statistic for each of the obtained gait decompositions.

	\subsection{Multi-Phase Gait Partitions Using NCD}
	\label{sec: partitioning}
	
	Using NCD, we can segment gait into a range of decompositions from kinematic information, where each decomposition is determined by a set of distinct transition conditions that are recurrent throughout the gait cycle. In this subsection, we report gait partitions obtained from the same data using NCD with different basis functions. Specifically, we use 20-second fragments of the collected walking data (refer to Section~\ref{sec: ekso_data} for details) for training our model and test the trained classifier on the trimmed dataset (a 1.33-minute time window). For each partitioning, we expand the state space through quadratic, cubic, and/or trigonometric functions of the original 12 states. Depending on the choice of basis functions, and consequently the recognized transition events, we segment gaits into 2, 4, and 6 phases, where in each case the phase transitions correspond to easily interpretable gait events. We test the segmentations on the collected dataset as explained in Section~\ref{sec: ekso_data}. We summarize our findings in Table~\ref{table: summary}.	
	
		\begin{figure}[!t]
		\centering
		\includegraphics[width=0.85\columnwidth]{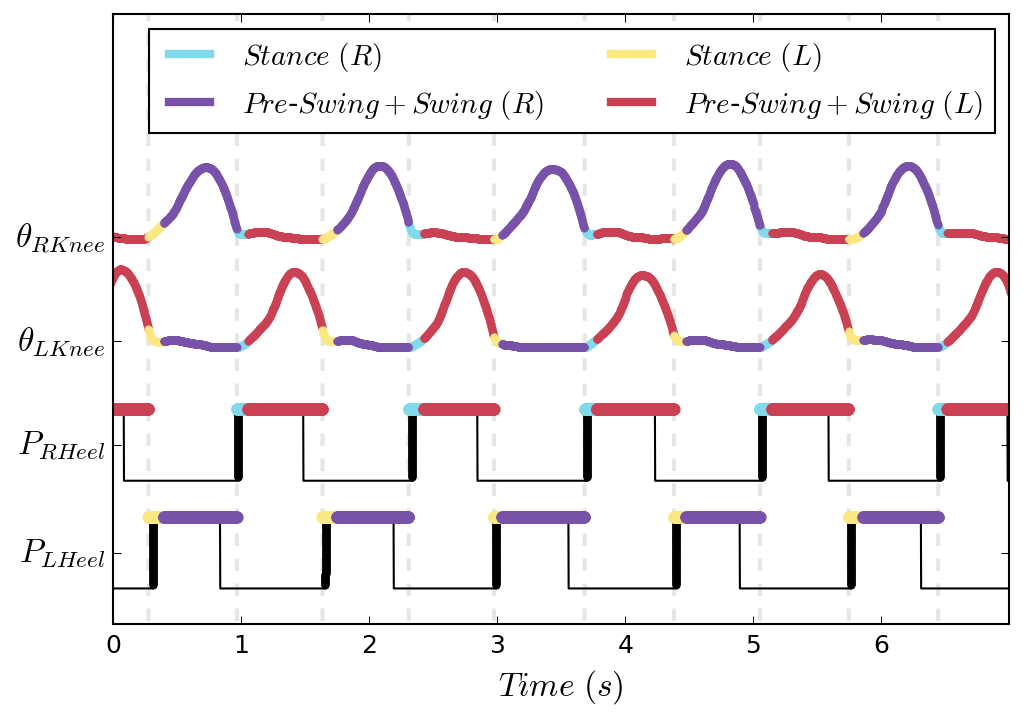}
		\vspace{-0.4cm}
		\caption{Four-phase segmentation of the gait cycle plotted against under-the-heel pressure sensors. Here, we are able to detect heel strike and the onset of knee buckling. We identify heel strikes with $5\%$ misclassifications and an average offset of $13.0ms \pm 8.0ms$. Note that pressure signals are not used in training for any decomposition.}
		\label{fig: 4phase_detection}
		\vspace{-0.3cm}
	\end{figure}
	
	\begin{figure*}[t]
		\centering
		\includegraphics[width=0.84\textwidth]{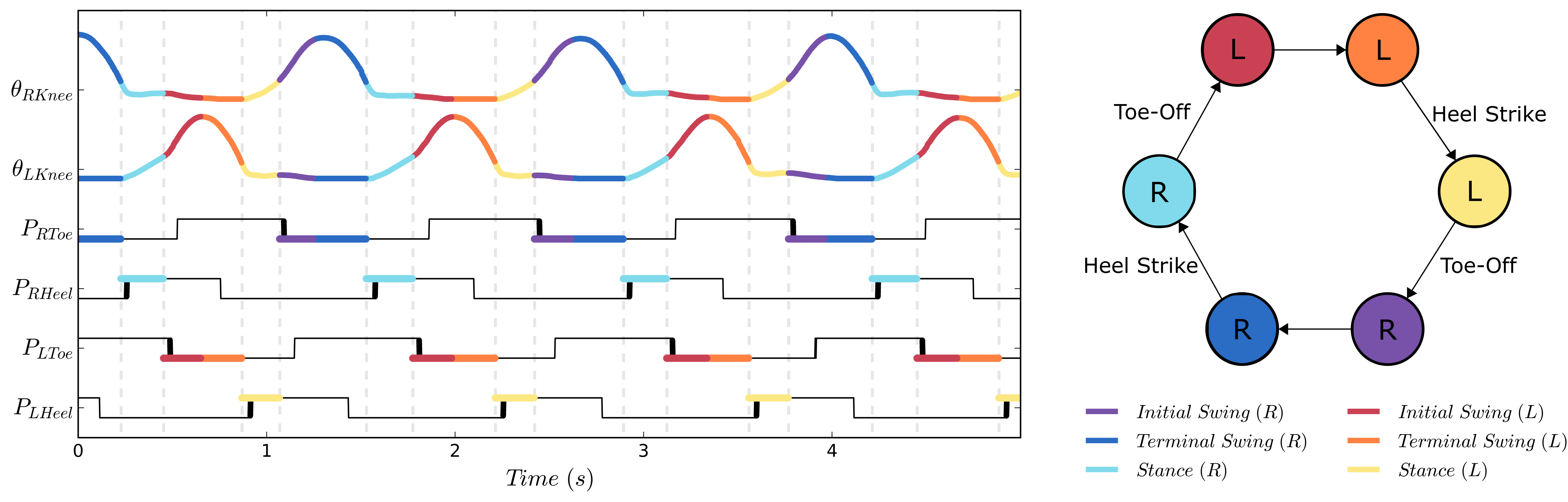}
		\vspace{-0.35cm}
		\caption{Six-phase segmentation of the gait cycle against foot-mounted pressure sensors. Dotted gray lines indicate NDC predictions of heel strikes and toe-offs, while the bold black segments indicate ground truth measurements. NCD reliably generated predictions of phase-transition events, misclassifying $9\%$ of the transitions and achieving an average absolute prediction offset of $20.0ms \pm 9.8ms$.}
		\label{fig: 6phase_detection}
		\vspace{-0.3cm}
	\end{figure*}

	In the 2-phase segmentation, we identify transitions that correspond to heel strikes. The gait cycle gets split into a right and left step, as shown in Fig.~\ref{fig: 2phase_detection}. We verify the accuracy of heel strike against pressure sensor signal and observe no misclassifications with a $\sim6.1ms$ offset. It is worth noting that all offsets were anticipatory, meaning that the predicted transitions occurred before the actual impact events.
	
	In the 4-phase partitioning, we identify transitions that correspond to heel strike and the onset of knee buckling, when the knee is ready to start bending but toe-off has not yet occurred. This decomposition, visible in Fig.~\ref{fig: 4phase_detection}, combines pre-swing and swing into one mode with a shortened stance as the second mode. Interestingly, this decomposition could be particularly relevant for control, because transition out of stance and into pre-swing can be interpreted as the cue for assistance, while heel strike (transition out of swing into stance) can be a signal for assistance to pause. Again, we verify the accuracy of heel strike predictions against heel strikes recorded via pressure sensors---results are reported in Table 1.  

	The 6-phase partitioning is the most complex of the ones reported here; we visualize it in Fig.~\ref{fig: 6phase_detection}. This partitioning allows us to identify both heel strikes and toe-offs, splitting the gait into right/left swing and stance. In addition, it divides swing into two segments of initial and terminal swing. Transition from initial into terminal swing corresponds to the clinically recognized foot clearance---when the swing leg passes the stance leg---and near-maximal knee flexion. This segmentation could also be extremely useful for assistance, because it allows us to detect the start of swing. In situations when we want the subject to independently initiate a step, we might want them to complete pre-swing without assistance and wait for toe-off to apply control. As before, we verify the toe-offs and heel strikes against pressure sensors---results are reported in Table 1.
	
	The list of possible gait partitions presented here is not exhaustive and is meant to illustrate the capabilities of the algorithm. Additional gait segmentations can be obtained, depending on what gait phase transitions are important in a particular application. 
	
	\section{Discussion and future work}
	We demonstrate that NCD in combination with a data-driven dynamics identification technique, such as the Koopman operator, can be used to infer mode transitions and hybrid dynamics of a system. We learn a data-driven model of a SLIP hopper and successfully use it in simulation to generate control. We further show the ability to complete gait partitioning for normal walking, where switching times for gait phases correspond to gait events measured independently through ground-reaction forces.  
	
	In future work, the learned partitioning of human gait could be used for closed-loop control of a lower-limb exoskeleton, similarily to how the learned model of the simulated SLIP hopper was used for control in Section~\ref{slip}. Specifically, the learned dynamics could be used to predict the system's forward evolution over time and to then calculate a stabilizing control signal using model predictive control to e.g., assist a person in an exoskeleton with balance. As mentioned in our results section, the classifier for human walking is imperfect---it experiences misclassifications and a $6$-$20~ms$ offset from ground truth during runtime. However, for a hardware implementation with closed-loop control, this performance could be further optimized. For instance, false positives are often caused by two or three mode transitions identified sequentially within $0.002~ms$ of each other---these could possibly be eliminated or lessened through the implementation of a high-pass filter. The reported offsets are always anticipatory, meaning that they consistently occur before the measured event, giving us the option to incorporate the average offset into our prediction of the timing of an impact event. Finally, the obtained gait event detection latencies are lower than those reported for human reaction times~\cite{walking_stumbling}. Thus, as is, the expected latencies should not be a limiting factor in safely controlling a dynamical system, e.g., an exoskeleton to avoid falling. 
	
	What is worth noting is that our method could be particularly well-suited for modeling abnormal gaits. For one, we do not parametrize the gait partitioning prior to application of NCD---we do not make assumptions about either the number of dynamic modes or phase durations. Using an unsupervised learning technique, we have the flexibility to identify any recurrent movement pattern and, as a result, partition gaits with varying granularity. We can identify motion patterns in individuals, even when the movements are not part of the ``correct" mode sequence in a specified task. Secondly, we do not require pre-labeling of phase transitions in the training data, which much prior work has relied on~\cite{mlp_joint_angles, lenating}, and are able to train our segmentation algorithm on small amounts of data (experiments in this study used 20-second sets of walking data sampled at $500Hz$). This means that gait partitions could be generated for individual patients and updated continually as their impairment changes over time. Personalized gait partitions from impaired individuals could then be used to generate tailored therapies or to create personalized control patterns for assistive devices. 
	
	Finally, the algorithm could further be used to identify higher level behaviors, such as stair climbing, marching, or walking down an incline. Through real-time identification of a person's activities based on kinematic data, NCD could facilitate providing task-relevant assistance without the need for manual task specification. Future work will seek experimental validation of our algorithms for generating assistance in lower-limb exoskeletons---both for flat-ground walking and for task-varying movement.
	
	\section{Conclusion}
	The presented data-driven methods are a step towards generating real-time model predictive controllers for hybrid systems, such as lower-limb exoskeletons. We show the ability to model and control a hybrid system---a 2D SLIP hopper---with no \textit{a priori} knowledge of the transition conditions or mode dynamics. We further show the ability to generate control-consistent gait partitions in a healthy individual. Long-term, our approach can lead to developing personalized models of gait for targeted assistance and/or rehabilitation in individuals with a range of walking impairments. 
    
	\addtolength{\textheight}{-0cm}   
	
	
	\bibliography{references}

\begin{thebibliography}{10}
\providecommand{\url}[1]{#1}
\csname url@samestyle\endcsname
\providecommand{\newblock}{\relax}
\providecommand{\bibinfo}[2]{#2}
\providecommand{\BIBentrySTDinterwordspacing}{\spaceskip=0pt\relax}
\providecommand{\BIBentryALTinterwordstretchfactor}{4}
\providecommand{\BIBentryALTinterwordspacing}{\spaceskip=\fontdimen2\font plus
\BIBentryALTinterwordstretchfactor\fontdimen3\font minus
  \fontdimen4\font\relax}
\providecommand{\BIBforeignlanguage}[2]{{%
\expandafter\ifx\csname l@#1\endcsname\relax
\typeout{** WARNING: IEEEtran.bst: No hyphenation pattern has been}%
\typeout{** loaded for the language `#1'. Using the pattern for}%
\typeout{** the default language instead.}%
\else
\language=\csname l@#1\endcsname
\fi
#2}}
\providecommand{\BIBdecl}{\relax}
\BIBdecl

\bibitem{koopman_ian}
I.~Abraham, G.~De~La~Torre, and T.~D. Murphey, ``Model-based control using
  {K}oopman operators,'' \emph{Proceedings of Robotics: Science and Systems
  (RSS)}, 2017.

\bibitem{deisenroth2015}
M.~P. Deisenroth, D.~Fox, and C.~E. Rasmussen, ``Gaussian processes for
  data-efficient learning in robotics and control,'' \emph{IEEE Transactions on
  Pattern Analysis and Machine Intelligence}, vol.~37, no.~2, pp. 408--423,
  2015.

\bibitem{pan2012}
Y.~Pan and J.~Wang, ``Model predictive control of unknown nonlinear dynamical
  systems based on recurrent neural networks,'' \emph{IEEE Transactions on
  Industrial Electronics}, vol.~59, no.~8, pp. 3089--3101, 2012.

\bibitem{mpc_neuralnets}
G.~Williams, N.~Wagener, B.~Goldfain, P.~Drews, J.~M. Rehg, B.~Boots, and E.~A.
  Theodorou, ``Information theoretic {MPC} for model-based reinforcement
  learning,'' in \emph{International Conference on Robotics and Automation
  (ICRA)}, 2017.

\bibitem{modellearning_review}
D.~Nguyen-Tuong and J.~Peters, ``Model learning for robot control: a survey,''
  \emph{Cognitive Processing}, 2011.

\bibitem{koopman_alex}
A.~Broad, T.~Murphey, and B.~Argall, ``Learning models for shared control of
  human-machine systems with unknown dynamics,'' \emph{Proceedings of Robotics:
  Science and Systems (RSS)}, 2018.

\bibitem{imu_neuralnet}
H.~T.~T. Vu, F.~Gomez, P.~Cherelle, D.~Lefeber, A.~Now{\'e}, and
  B.~Vanderborght, ``{ED}-{FNN}: A new deep learning algorithm to detect
  percentage of the gait cycle for powered prostheses,'' \emph{Sensors},
  vol.~18, no.~7, p. 2389, 2018.

\bibitem{mlp_joint_angles}
C.~Chen, D.~Liu, X.~Wang, C.~Wang, and X.~Wu, ``An adaptive gait learning
  strategy for lower limb exoskeleton robot,'' in \emph{IEEE International
  Conference on Real-time Computing and Robotics (RCAR)}, 2017, pp. 172--177.

\bibitem{HMM_gaitpartitioning}
J.~Taborri, S.~Rossi, E.~Palermo, F.~Patan{\`e}, and P.~Cappa, ``A novel {HMM}
  distributed classifier for the detection of gait phases by means of a
  wearable inertial sensor network,'' \emph{Sensors}, vol.~14, no.~9, pp.
  16\,212--16\,234, 2014.

\bibitem{GMM_gaitpartitioning}
J.-U. Chu, K.-I. Song, S.~Han, S.~H. Lee, J.~Y. Kang, D.~Hwang, J.-K.~F. Suh,
  K.~Choi, and I.~Youn, ``Gait phase detection from sciatic nerve recordings in
  functional electrical stimulation systems for foot drop correction,''
  \emph{Physiological Measurement}, 2013.

\bibitem{henzinger1996}
T.~A. Henzinger, ``The theory of hybrid automata,'' in \emph{Proceedings of the
  IEEE Symposium on Logic in Computer Science}, 1996, pp. 278--283.

\bibitem{williams2016}
M.~Williams, I.~Kevrekidis, and C.~Rowley, ``A data--driven approximation of
  the {K}oopman operator: Extending dynamic mode decomposition,'' \emph{Journal
  of Nonlinear Science}, vol.~25, pp. 1307--1346, 2015.

\bibitem{hdbscan}
R.~Campello, D.~Moulavi, and J.~Sander, ``Density-based clustering based on
  hierarchical density estimates,'' in \emph{Advances in Knowledge Discovery
  and Data Mining}.\hskip 1em plus 0.5em minus 0.4em\relax Springer, 2013, pp.
  160--172.

\bibitem{machinelearning}
C.~Bishop, \emph{Pattern Recognition and Machine Learning}.\hskip 1em plus
  0.5em minus 0.4em\relax Springer, 2006.

\bibitem{koopman}
B.~Koopman, ``Hamiltonian systems and transformation in {H}ilbert space,''
  \emph{Proceedings of the National Academy of Sciences}, vol.~17, no.~5, pp.
  315--318, 1931.

\bibitem{budisic2012}
M.~Budi\v{s}i\'{c}, R.~Mohr, and I.~Mezi\'{c}, ``Applied {K}oopmanism,''
  \emph{Chaos: An Interdisciplinary Journal of Nonlinear Science}, vol.~22,
  no.~4, p. 047510, 2012.

\bibitem{ola_RSS}
A.~Kalinowska, K.~Fitzsimons, J.~Dewald, and T.~D. Murphey, ``Online user
  assessment for minimal intervention during task-based robotic assistance,''
  \emph{Proceedings of Robotics: Science and Systems (RSS)}, 2018.

\bibitem{alex_SAC}
A.~R. Ansari and T.~D. Murphey, ``Sequential action control: closed-form
  optimal control for nonlinear and nonsmooth systems,'' \emph{IEEE
  Transactions on Robotics}, vol.~32, no.~5, pp. 1196--1214, 2016.

\bibitem{gaitanalysis}
H.~G. Chambers and D.~H. Sutherland, ``A practical guide to gait analysis,''
  \emph{Journal of the American Academy of Orthopaedic Surgeons (JAAOS)},
  vol.~10, no.~3, pp. 222--231, 2002.

\bibitem{walking_stumbling}
A.~Schillings, B.~Van~Wezel, and J.~Duysens, ``Mechanically induced stumbling
  during human treadmill walking,'' \emph{Journal of neuroscience methods},
  vol.~67, no.~1, pp. 11--17, 1996.

\bibitem{lenating}
L.~Drnach, I.~Essa, and L.~H. Ting, ``Identifying gait phases from joint
  kinematics during walking with switched linear dynamical systems,'' in
  \emph{IEEE International Conference on Biomedical Robotics and
  Biomechatronics (BioRob)}, 2018.

\end{thebibliography}

\end{document}